\documentclass[11pt,a4paper]{article}

\usepackage[margin=1in]{geometry}  
\usepackage{titlesec}  
\usepackage{times}  
\usepackage{graphicx}  
\usepackage{hyperref}  

\titleformat{\section}
  {\normalfont\large\bfseries}{\thesection}{1em}{}
\titleformat{\subsection}
  {\normalfont\normalsize\bfseries}{\thesubsection}{1em}{}

\usepackage{graphicx}
\usepackage{amsmath}
\usepackage{algorithm}
\usepackage{algpseudocode}
\usepackage{hyperref}
\usepackage[dvipsnames]{xcolor}
\usepackage{authblk}

\usepackage{listings}
\lstdefinestyle{prompt}{
    breaklines=true,
    postbreak=\raisebox{0ex}[0ex][0ex]{\ensuremath{\hookrightarrow\space}},
    basicstyle=\small\ttfamily,
    frame=single,
    framesep=5pt,
    numberstyle=\tiny,
    numbers=left,
    captionpos=b,
    backgroundcolor=\color{gray!10},
    commentstyle=\color{gray!60}
}

\begin{document}

\title{Benchmarking Floworks against OpenAI \& Anthropic: A Novel Framework for Enhanced LLM Function Calling}

\author[1]{Nirav Bhan\thanks{Lead and Corresponding author}}
\author[1]{Shival Gupta\thanks{Lead author}}
\author[1]{Sai Manaswini}
\author[1,2]{Ritik Baba}
\author[1]{Narun Yadav}
\author[1]{Hillori Desai}
\author[1,3]{Yash Choudhary}
\author[1]{Aman Pawar}
\author[1]{Sarthak Shrivastava}
\author[1]{Sudipta Biswas}

\affil[1]{Floworks}
\affil[2]{Indian Institute of Technology Kharagpur}
\affil[3]{Indian Institute of Technology Bombay}

\renewcommand\Authands{ and }



\maketitle

\begin{abstract}
Large Language Models (LLMs) have shown remarkable capabilities in various domains, yet their economic impact has been limited by challenges in tool use and function calling. This paper introduces ThorV2, a novel architecture that significantly enhances LLMs' function calling abilities. We develop a comprehensive benchmark focused on HubSpot CRM operations to evaluate ThorV2 against leading models from OpenAI and Anthropic. Our results demonstrate that ThorV2 outperforms existing models in accuracy, reliability, latency, and cost efficiency for both single and multi-API calling tasks. We also show that ThorV2 is far more reliable and scales better to multi-step tasks compared to traditional models. Our work offers the tantalizing possibility of more accurate function-calling compared to today's best-performing models using significantly smaller LLMs. These advancements have significant implications for the development of more capable AI assistants and the broader application of LLMs in real-world scenarios.
\end{abstract}

\section{Introduction}
Large Language Models (LLMs) have revolutionized natural language processing and artificial intelligence, demonstrating remarkable capabilities across a wide range of tasks (\cite{naveed2023comprehensive, hadi2023large}). However, their economic impact has been somewhat limited, particularly in domains requiring precise interaction with external tools and APIs. A key challenge in this area is the task of function calling, where LLMs must accurately interpret user queries and translate them into appropriate API calls.

The underwhelming performance of LLMs in function calling has had tangible consequences. Much-anticipated AI gadgets like RabbitR1 and Humane AI Pin have faced criticism due to their inability to reliably fulfill user tasks. OpenAI's GPT-store has similarly met with a very lukewarm response. We also note that while chatbots and coding assistants can increase productivity, almost no jobs have been completely taken over by AI as of September 2024 \cite{seoai_ai_jobs_2024}. This highlights a critical gap between the theoretical capabilities of LLMs and their practical performance in the real world.

To address these challenges, we introduce ThorV2, a novel architecture designed to enhance LLMs' function calling abilities. ThorV2 employs an innovative approach we term "edge-of-domain modeling," which focuses on correcting errors rather than providing comprehensive instructions upfront. This paper presents a thorough evaluation of ThorV2 against leading models from OpenAI and Anthropic, using a new benchmark based on HubSpot CRM operations.

Our contributions in this paper are as follows:
\begin{itemize}
    \item We introduce ThorV2, a novel architecture for enhancing LLM function calling capabilities.
    \item We develop a comprehensive benchmark for evaluating function calling performance in the context of CRM operations.
    \item We propose a new metric, Reliability, to measure consistent performance across repeated tests.
    \item We demonstrate ThorV2's superior performance in accuracy, reliability, latency, and cost efficiency compared to leading commercial models.
    \item We show ThorV2's ability to generalize to complex, multi-step tasks with minimal performance degradation.
\end{itemize}

The rest of this paper is organized as follows: Section 2 discusses related work and the limitations of existing approaches. Section 3 describes the ThorV2 architecture in detail. Section 4 outlines our evaluation methodology, including the benchmark dataset and metrics. Section 5 presents our results and analysis. Section 6 discusses the implications of our findings and potential for generalization. Finally, Section 7 concludes the paper and suggests directions for future work.

\section{Limitations of Traditional Function Calling}

Traditional AI systems have treated function calling as a monolithic task, where the model accepts a task and relevant function schemas, then outputs a complete function call. This approach suffers from several key disadvantages:

\begin{itemize}
    \item \textbf{Inefficient Function Retrieval:} Retrieving appropriate functions often relies on vector similarity, a heuristic approach known to suffer from issues with accuracy, scalability, and domain specificity, as discussed in \cite{zhou2022problems}.
    
    \item \textbf{Excessive Token Lengths:} Function schemas can be lengthy, leading to large prompt sizes. This increases deployment costs, time consumption, and can result in decreased accuracy on reasoning tasks. For example, \cite{levy2024same} shows that reasoning abilities of LLMs fall drastically with increase in active context length.
    
    \item \textbf{High Output Sensitivity:} LLMs are trained on free-flowing text and struggle with the rigid requirements of function calling, where precise variable names, JSON structures, and argument values are crucial.
\end{itemize}

These limitations have resulted in even the best closed-source LLMs (e.g., GPT-4o, Claude-3 Opus) failing to solve the function calling problem effectively.

\section{ThorV2 Architecture}
\subsection{Overview of ThorV2}
ThorV2 is a Cognitive Enhancement Architecture (CEA) designed to augment the function calling capabilities of Large Language Models. Unlike traditional approaches that rely on comprehensive upfront instructions, ThorV2 employs a novel strategy we term ``edge-of-domain modeling". This approach focuses on identifying and correcting errors in the LLM's output, rather than attempting to preemptively cover all possible scenarios. We propose a new type of framework called ``Agent-Validator architecture" which puts this type of modeling into practice. Later, we also discuss Composite Planning and other architectural optimizations to improve the cognitive performance of LLMs.

\subsection{Edge-of-domain Modeling}

To build intuition for ThorV2's architecture, we first discuss a general principle on providing instructions for any task. One approach can be to provide the entire knowledge base upfront, hoping to cover all possible cases that an agent could encounter during the execution of the task. The other approach would be to only provide minimal instruction upfront, allowing the agent to begin the task, and provide the remaining information either in the form of just-in-time instructions or post-task error-correction.

\begin{figure}[h]
\centering
\includegraphics[width=0.9\linewidth]{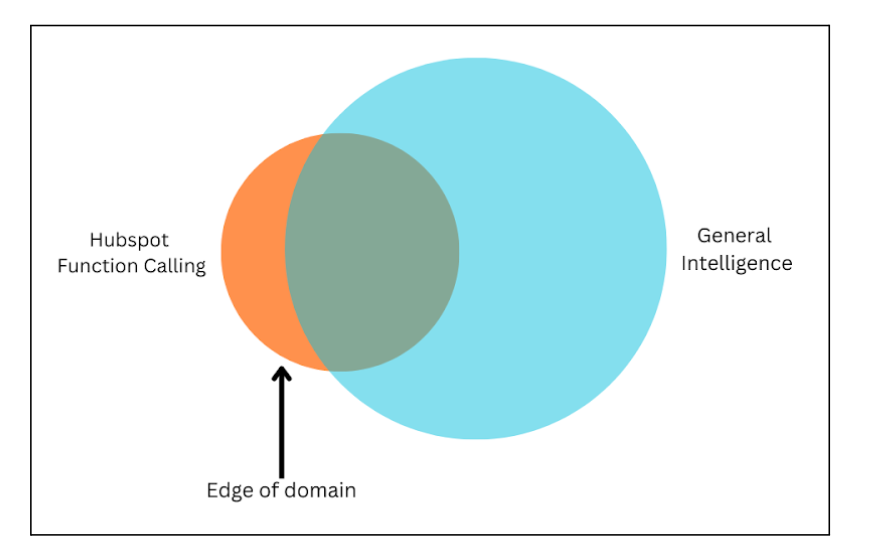}
\caption{Edge-of-domain modeling}
\label{fig:edge_of_domain}
\end{figure}

\textbf{\textit{Edge-of-domain modeling}} refers to providing instructions through iteration and error-correction. We contrast this with the traditional approach of providing the entire world model in the system prompt before beginning, hoping to accomplish the task in a single attempt. We call the latter approach \textbf{\textit{whole-of-domain modeling}}. The 2 approaches are contrasted in figure \ref{fig:edge_of_domain}.

In the figure, the orange disc represents the full knowledge set required to perform Function Calling on Hubspot, while the light blue disc represents the knowledge already possessed by the model. With whole-of-domain modeling, the entire orange disc needs to be provided to the model, whereas with edge-of-domain modeling, the difference between the orange and blue spheres is sufficient. This significantly reduces the information load on the LLM. This approach has several advantages:

\begin{itemize}
    \item \textbf{Reduced token count}: By eliminating the need for extensive upfront instructions, we dramatically reduce the number of tokens in the prompt. This leads to improved accuracy, reduced cost, and lower latency.
    \item \textbf{Improved scalability}: The error correction approach scales more efficiently to complex, multi-step tasks compared to traditional instruction-based methods.
    \item \textbf{Enhanced reliability}: By consistently correcting specific types of errors, ThorV2 achieves a high degree of reliability in its outputs.
\end{itemize}


\subsection{Agent-Validator Architecture}

\begin{figure*}[h]
\centering
\fbox{\includegraphics[width=0.9\linewidth]{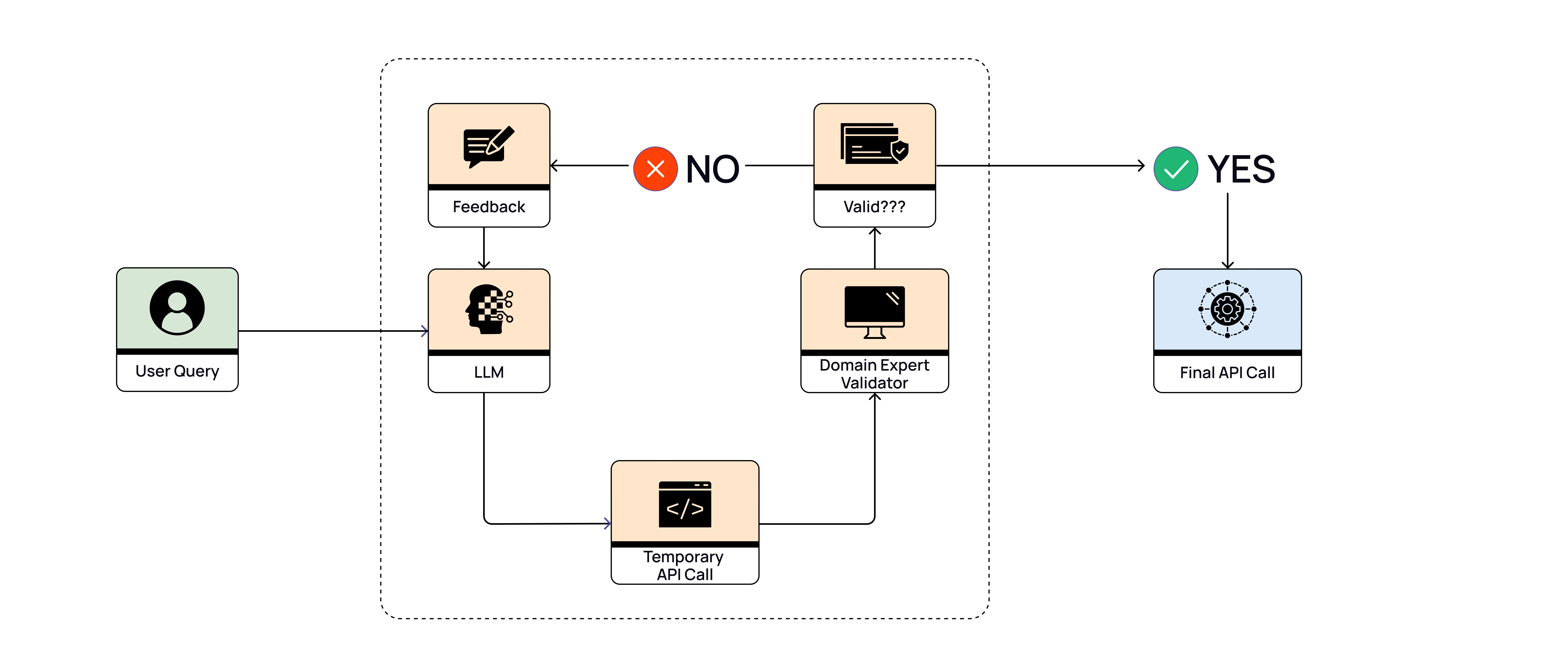}}
\caption{Thor's Agent-Validator architecture}
\label{fig:thor_validator}
\end{figure*}

This edge-of-domain modeling is implemented through an Agent-Validator Architecture, as shown in figure \ref{fig:thor_validator}. They key component of this system are Domain Expert Validators (DEVs), static agents written entirely in code that inspect the API calls generated by the LLM for errors. If an error is detected, the DEV provides corrective feedback to the LLM, allowing it to remedy the mistake. This iterative process continues until a correct API call is generated or a maximum number of attempts is reached. The name "domain expert validator" underscores the fact that these are highly specialized modules, possessing significant domain knowledge.

\subsubsection{Validator example}

While the full details of our Validator are proprietary, we give an example here to illustrate the general idea. The HubSpot API has two kinds of retrieval queries: retrieving properties which can be done in a single API call, and retrieving associations which require two API calls. For example, a deal's \textbf{\textit{amount}} and \textbf{\textit{closing date}} are properties, but \textbf{\textit{notes}} are separate objects associated with the deal.  LLMs frequently confuse these two operations, wrongly treating associations as properties. Our Validator identifies this error and provides the following feedback to the LLM: \textbf{``Hubspot needs you to search for associated resource first and use its deal id as the associated resource id in your second query. Break into two steps and do variable injection"}. If the LLM makes the same error again, the Validator will keep repeating its feedback until the error is fixed -- hence the loop in figure \ref{fig:thor_validator}.

\subsubsection{Relation to other approaches}

We contrast our approach with another set of approaches called ``Agent-Critic Systems" or ``Agentic Workflows". In these systems, it is common to have a primary LLM ``agent" receive feedback from other LLMs who act as ``critics" (\cite{bai2022constitutional, qian2024chatdev, wu2023autogen}). These additional LLMs not only increase time and cost, but suffer from high error rates defeating the very purpose of a critic. In other words, we do not believe LLMs are broadly capable of self-critique, which is exemplified by the limited accuracy of agentic systems in benchmarks like SWE-Bench\footnote{As per the SWE-bench on 5th September 2024, the highest accuracy on Full-SWE-bench is 22\%. \url{https://www.swebench.com/}}. as well as real-world tasks. (\cite{guo2024large, jimenez2024swebench}). 

\textbf{\textit{Our approach differs from these in that we employ an entirely static critic}}. Static critics require domain knowledge and some engineering effort to develop, but they have the advantage of being extremely fast and perfectly accurate. The use of static DEVs is possible because the errors made by LLMs in function calling tend to be highly repetitive and predictable. Our approach bears some resemblance to the LLM-modulo approaches proposed in \cite{kambhampati2024llms}. However, rather than rely on external verifiers that must cover every possible instance, we develop our own verifiers to correct the most common errors committed by the LLM agent. By focusing on these common error patterns, we can achieve significant improvements in accuracy and reliability.

\subsection{Composite Planning Approach}
\label{sec:composite_planning}

For multi-step tasks, ThorV2 employs a novel \textit{composite planning} approach. This allows for the generation of multiple API calls in a single step, instead of the traditional planning approach which generates API calls one at a time.

\begin{figure*}[!htbp]
\centering
\includegraphics[width=0.9\linewidth]{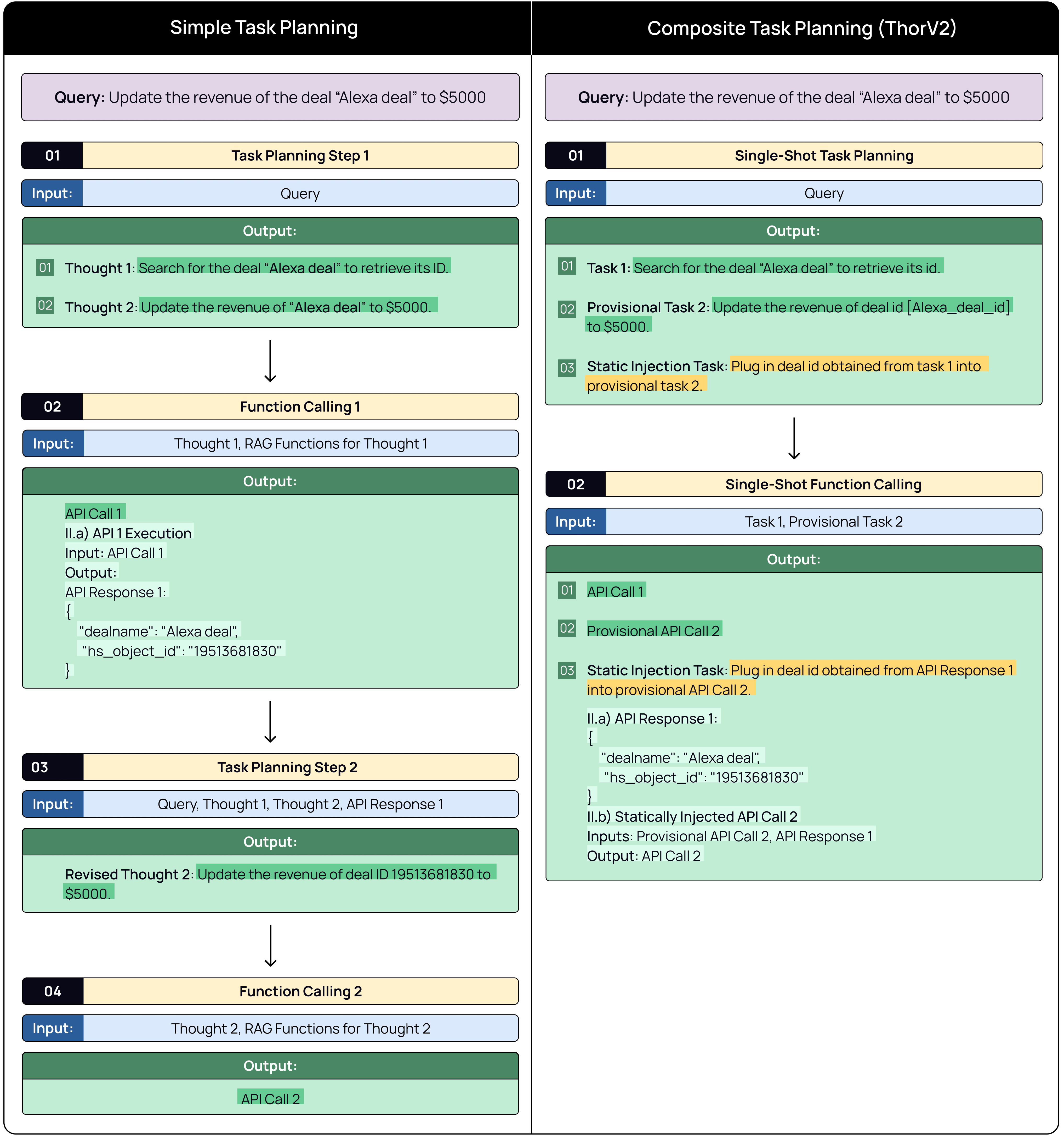}
\caption{Composite Planning Approach}
\label{fig:composite_planning}
\end{figure*}

The key idea is to use placeholder variables to represent unknown values, and a planning logic to inject the values automatically. 
Once an API response is obtained, the values of placeholder variables are injected into the next API call using a combination of agents and code. A detailed walkthrough comparing this approach with traditional planning is provided in figure \ref{fig:composite_planning}. Generating multiple API calls at once requires sophisticated planning and reasoning capabilities, which is very challenging for ordinary LLMs. Our Agent-Validator architecture simplifies this process as well by correcting errors in the planning step. 


Our novel approach offers several benefits:

\begin{itemize}
    \item \textbf{Reduced latency}: By generating multiple API calls at once, ThorV2 minimizes the back-and-forth communication typically required in sequential planning.
    \item \textbf{Improved efficiency}: The use of placeholder variables allows for more efficient handling of interdependent tasks.
    \item \textbf{Better scaling}: Composite planning results in sublinear growth in latency as task complexity increases, in contrast to the superlinear growth observed in traditional models.
\end{itemize}

These abilities make ThorV2 a powerful tool for tackling multi-step queries. We capture their impact through scaling laws which are discussed in section \ref{sec:scaling_laws}.

\subsection{Other architectural optimizations}

Some of our other optimizations include using a token-efficient information-preserving intermediate language for generating the API call. We find in practice that not only does this reduce cost and generation time, but it also increased model accuracy. It appears that LLMs, being trained on text which is mostly natural language, have an innate aversion to writing API calls in JSON format, and instead prefer to output something that looks more like English text. A similar result was found in \cite{tam2024let}, showing that forcing a particular structure on an LLM's output has the effect of reducing its intelligence on its primary task.

We also simplify all deterministic components of the API call generation task, performing them using a static module, thereby reducing the task complexity and increasing accuracy for the LLM. In some cases, the API call generation requires another task which is of comparable complexity such as a complicated math operation. However, the complicated operation cannot be done statically. In this case, we use a separate LLM-based agent to perform this task. This reduces the intelligence burden for the primary LLM, boosting the overall function calling accuracy.

\section{Evaluation Methodology}
\subsection{Benchmark Dataset}
To evaluate the performance of ThorV2, we developed a comprehensive benchmark called \textbf{HubBench}\footnote{This dataset is public and can be found at: https://huggingface.co/datasets/floworks/HubBench-queries} based on HubSpot CRM operations.

Hubspot is a CRM (Customer Relationship Management) application that allows easy management of customer and sales databases. It is one of the most popular CRM applications used by sales teams around the world. We access this tool using the Hubspot REST API, which supports 5 types of elemental operations:  Create, Read, Update, Delete, and Associate (CRUDA) operations. The evaluation dataset consists of:

\begin{itemize}
    \item 142 single-API queries covering CRUDA operations
    \item 58 complex queries requiring exactly two API calls to execute.
\end{itemize}

\begin{figure*}[h]
\centering
\includegraphics[width=0.8\linewidth]{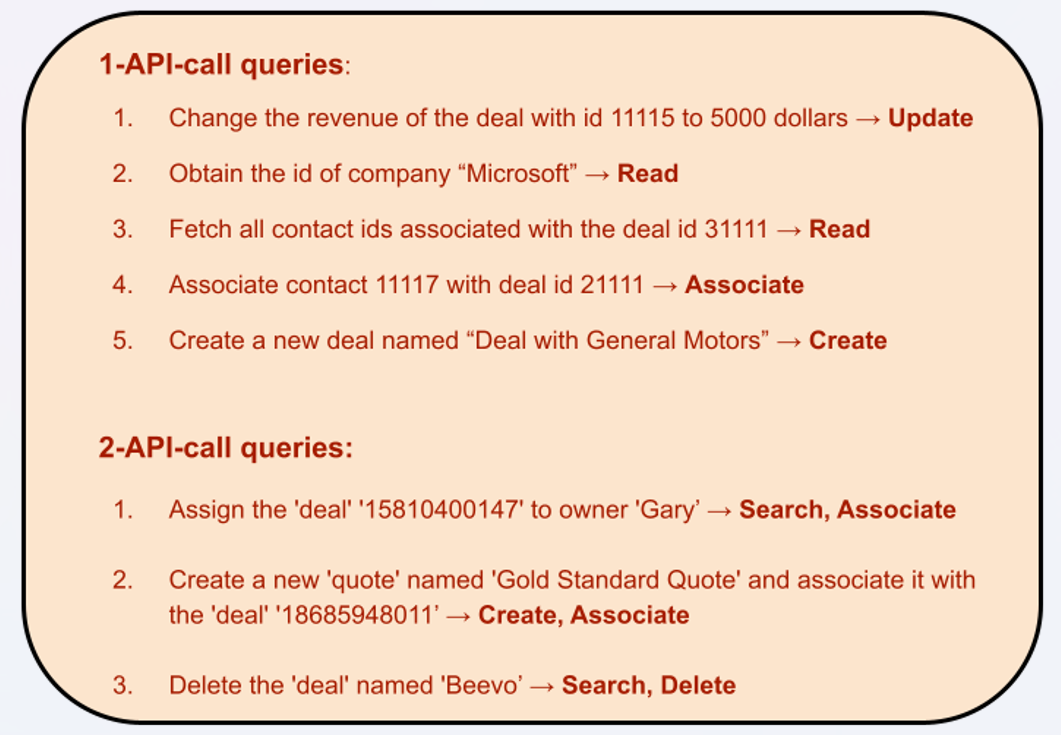}
\caption{HubBench Sample Queries}
\label{fig:hubbench_queries}
\end{figure*}

Some sample queries from this dataset are shown in figure \ref{fig:hubbench_queries}. These queries have been created to represent a wide range of real-world CRM tasks, ensuring a thorough evaluation of function calling capabilities.

\subsection{Evaluation Metrics}
We assess the performance of the models using four key metrics:

\begin{itemize}
    \item Accuracy: The percentage of queries for which a perfect API call is generated.
    \item Reliability: The percentage of queries for which the API success response remains consistent over 10 repeated attempts.
    \item Latency: The total execution time from query submission to API call generation.
    \item Cost: The monetary cost of generating the API call, based on the pricing of proprietary LLM APIs.
\end{itemize}

\subsection{Accuracy Measurement}
Accuracy is a crucial metric in our evaluation, defined as the percentage of queries resulting in a perfect API call generation. A perfect API call must satisfy two conditions: it must be executable (i.e. well-formed) and it must fulfill the intended task. Our accuracy assessment employs a rigorous two-stage process:

\begin{enumerate}
    \item \textbf{Software Evaluation:} We verify whether the generated API call executes without error on the Hubspot API.
    \item \textbf{Human Evaluation:} Similar to the authors of \cite{zeng2023evaluating}, we find that LLMs are not great at evaluating function calling. Therefore, we add a second layer of blind human evaluation, assessing five key criteria:
    \begin{itemize}
        \item Function Selection: Appropriateness of the chosen function for the query.
        \item Task Representation: Accurate expression of the user's request in the API call.
        \item Structural Integrity: Correctness and consistency of the API call structure with the Schema.
        \item Functional Integrity: Provision of all essential arguments with appropriate values.
        \item Instruction Containment: Absence of extraneous harmful or undesired operations.
    \end{itemize}
\end{enumerate}

An API call is considered correct only if it satisfies all five human evaluation criteria and passes the software evaluation. This comprehensive approach ensures a thorough assessment of the model's accuracy in generating API calls.

\subsection{Reliability Measurement}
Reliability in the context of Large Language Models (LLMs) for function calling is particularly crucial, given the inherent non-deterministic nature of these models \cite{ouyang2023llm}. While this non-determinism can be beneficial for creative tasks, it poses challenges in function calling scenarios where consistency is paramount. In this paper, we define reliability as the consistency of the model's performance across multiple attempts. Our measurement approach is as follows:

\begin{enumerate}
    \item We execute the entire test suite 10 times.
    \item We identify ``fluctuating queries" -- queries that yield at least one Pass and one Fail across the 10 attempts. Queries that consistently yield the same result (all Pass or all Fail) are categorized as non-fluctuating or ``consistent".
    \item We calculate the reliability metric using the following formula:
    
    \[ \text{Reliability} = \frac{\text{Number of Consistent Queries}}{\text{Total Queries in Test Suite}} \times 100\% \]
\end{enumerate}

It is important to note that reliability differs from accuracy. A perfectly reliable model (100\% reliability) doesn't necessarily achieve perfect accuracy; rather, it consistently produces the same results, whether correct or incorrect, across multiple runs. This reliability metric allows us to assess the model's consistency in API call generation, a critical factor in building dependable AI-driven systems for real-world applications.

\subsection{Comparison Models}
We compare ThorV2 against three state-of-the-art commercial models from leading AI firms like OpenAI and Anthropic:

\begin{itemize}
    \item Claude-3 Opus (Anthropic)
    \item GPT-4o \footnote{Specifically, GPT-4o-2024-05-13} (OpenAI) 
    \item GPT-4-turbo \footnote{Specifically, GPT-4-turbo-2024-04-09} (OpenAI) 
\end{itemize}

These models were chosen because they were widely considered to be the best available models as of May 2024, and also because they are explicitly provided with function calling APIs. They  represent the current benchmark for LLM performance and provide a robust baseline for evaluating ThorV2's capabilities. These models are hereafter referred to as "Comparison Models".

\subsection{Comparison Model Architecture}

To provide a fair comparison, we implemented specific architectures for both single and multi-API function calling tasks for the comparison models (Claude-3 Opus, GPT-4o, and GPT-4-turbo). These architectures are designed to leverage the models' capabilities while adhering to their standard usage patterns. It is important to note that while ThorV2 due to its versatile nature does not require separate implementations for multi-API function calling, for comparison models it is essential to provide a different architecture which explicitly takes care of planning and step-by-step execution. Without these changes, the performance of Comparison models would lag way behind ( $\leq 15 \%$ accuracy in our tests).  

\subsubsection{Single API Function Calling Architecture}

For single API function calling, we use a straightforward approach that mimics typical usage of these models for function calling tasks. This approach uses Retrieval Augmented Generation (RAG) to provide the 5 most relevant functions to the LLM. Since the goal of our study is to focus on the LLM's capabilities, we have set up the Benchmark in such a way that the 5 functions given to the LLM \textit{always} contain the correct function for solving the  query. 

\begin{figure}[h]
\centering
\fbox{\includegraphics[width=0.6\linewidth]{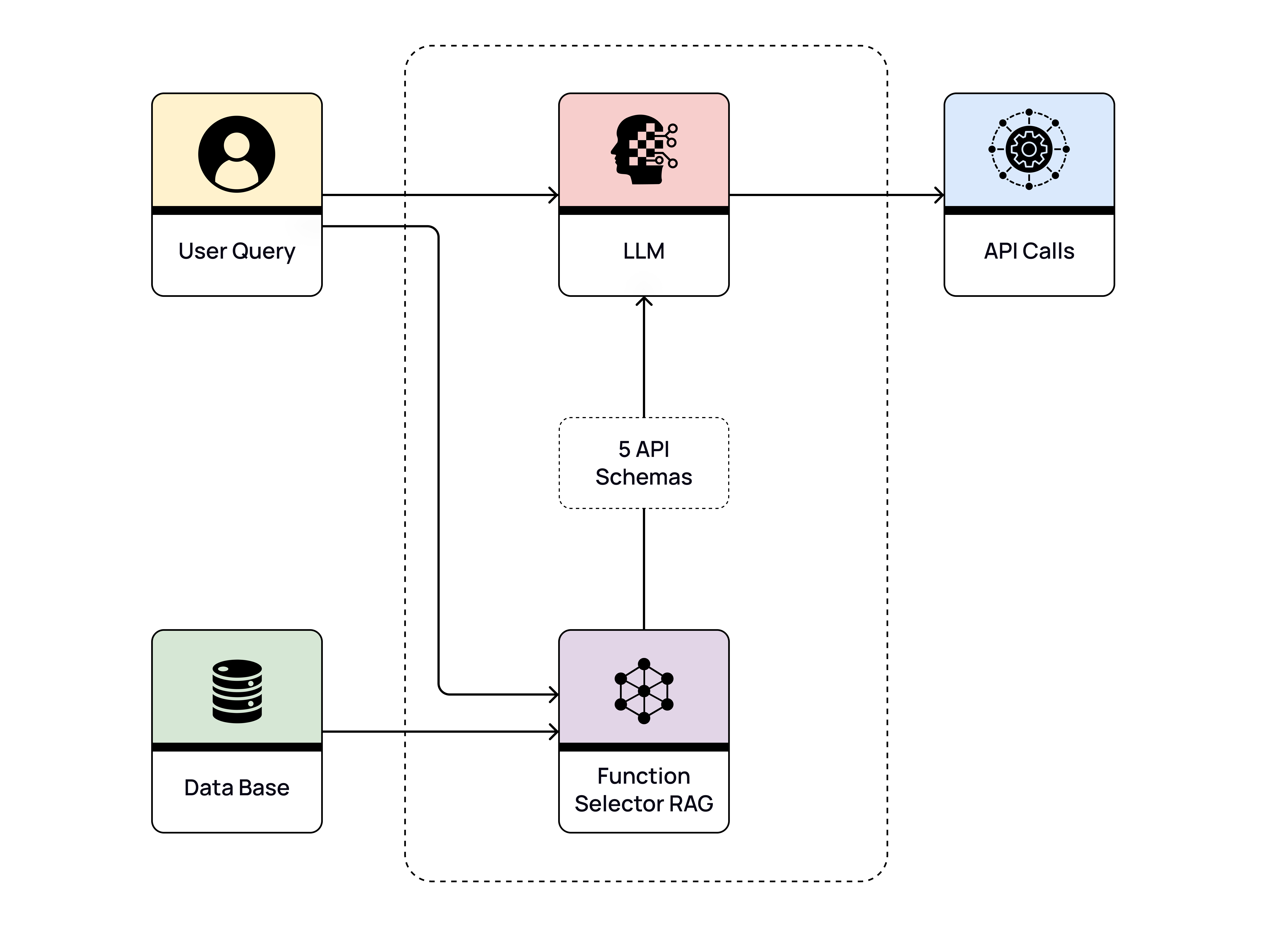}}
\caption{Single API Function Calling Architecture for Comparison Models}
\label{fig:single_api_architecture}
\end{figure}

The following steps are undertaken for function calling:
\begin{enumerate}
    \item The user query is provided as input to the entire system
    \item The RAG module fetches 5 relevant functions from its database for solving the given query, one of which is always correct
    \item The user query and 5 relevant API schemas are provided to the LLM
    \item The LLM processes the input and generates an API call (via the function-calling API)
    \item The API call is sent to the HubSpot API, and the response is returned
\end{enumerate}

This approach allows the models to directly translate user queries into API calls based on the provided schemas, simulating real-world function calling scenarios.

\subsubsection{Multi-API Function Calling Architecture}

For multi-API function calling, we implement a more complex architecture to handle interdependent API calls.

\begin{figure}[h]
\centering
\fbox{\includegraphics[width=0.8\linewidth]{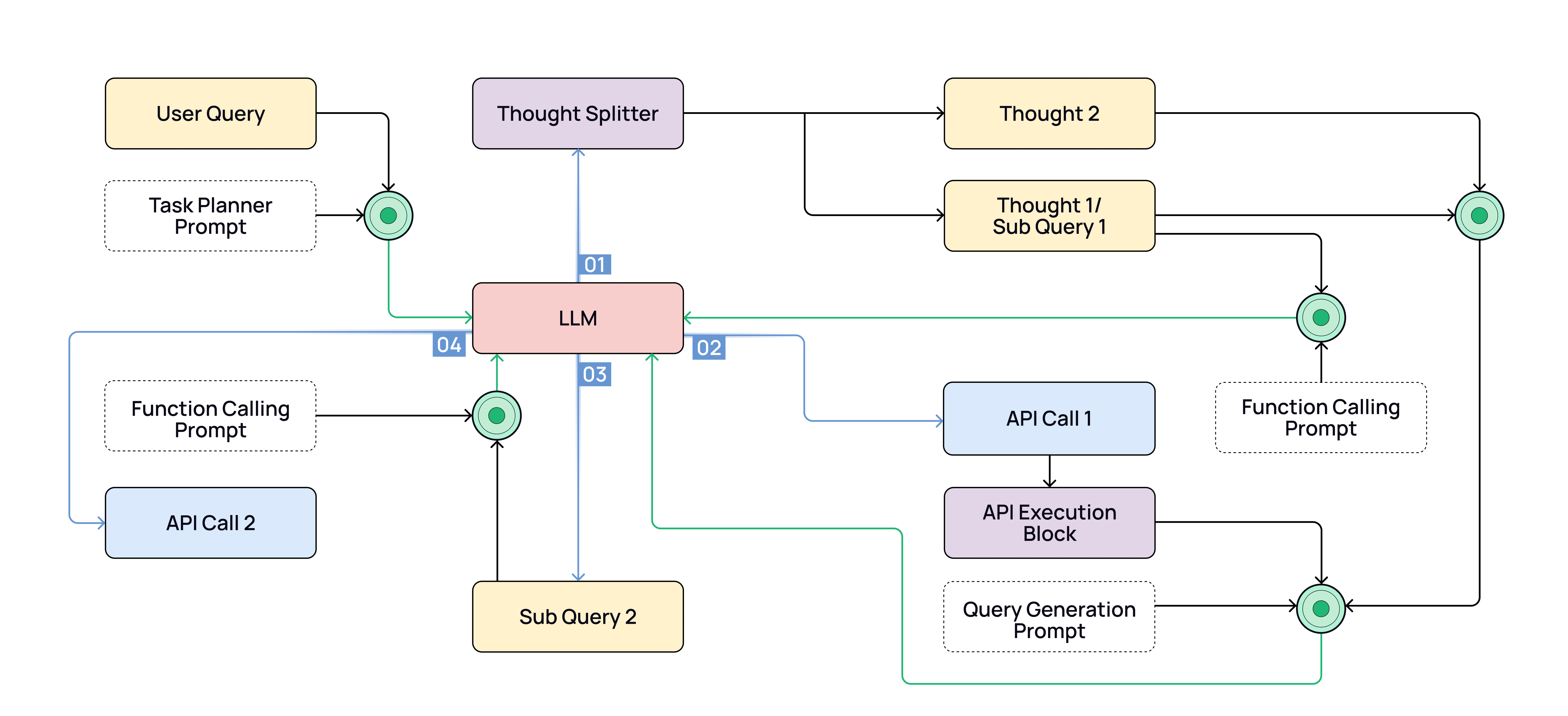}}
\caption{Multi-API Function Calling Architecture for Comparison Models}
\label{fig:multi_api_architecture}
\end{figure}

This architecture consists of the following steps:
\begin{enumerate}
    \item The LLM splits the input query into two tasks according to a predefined format.
    \item A RAG (Retrieval-Augmented Generation) block fetches the nearest 5 API schemas for each subquery.
    \item The LLM executes the first task, generating the first API call.
    \item An API execution block sends the first API call to the Hubspot endpoint and receives the response.
    \item The LLM uses the API response to modify the second task.
    \item The LLM executes the second task, generating the second API call.
    \item The final API call is executed, and the response is returned.
\end{enumerate}

This architecture allows the comparison models to handle complex, multi-step queries by breaking them down into manageable subtasks. It also uses the intermediate results to inform subsequent API calls.

Both architectures are designed to maximize the performance of the comparison models within the constraints of their standard usage patterns, ensuring a fair comparison with our ThorV2 system.

\section{Results and Analysis}
\subsection{Single API Calling Performance}
We evaluated ThorV2 against three state-of-the-art models: Claude-3 Opus, GPT-4o, and GPT-4-turbo. The results for single API calling tasks are summarized in Table \ref{tab:single_api_results}.

\begin{table}[h]
\centering
\caption{Performance metrics for single API calling tasks}
\label{tab:single_api_results}
\vspace{0.7em}
\begin{tabular}{|p{0.23\textwidth}|p{0.12\textwidth}|p{0.12\textwidth}|p{0.12\textwidth}|p{0.12\textwidth}|}
\hline
Model & Accuracy & Reliability & Latency (s) & Cost (\$/1000 queries) \\
\hline
Floworks-ThorV2 & \textbf{90.1\%} & \textbf{100\%} & \textbf{2.29} & \textbf{\$1.60} \\
Claude-3 Opus & 78.2\% & 59.7\% & 15.3 & \$46.7 \\
GPT-4o & 51.4\% & 83.9\% & 2.92 & \$4.14 \\
GPT-4-turbo & 48.6\% & 86.6\% & 4.55 & \$6.15 \\
\hline
\end{tabular}
\end{table}


\begin{figure}[h]
\centering
\fbox{\includegraphics[width=0.8\linewidth]{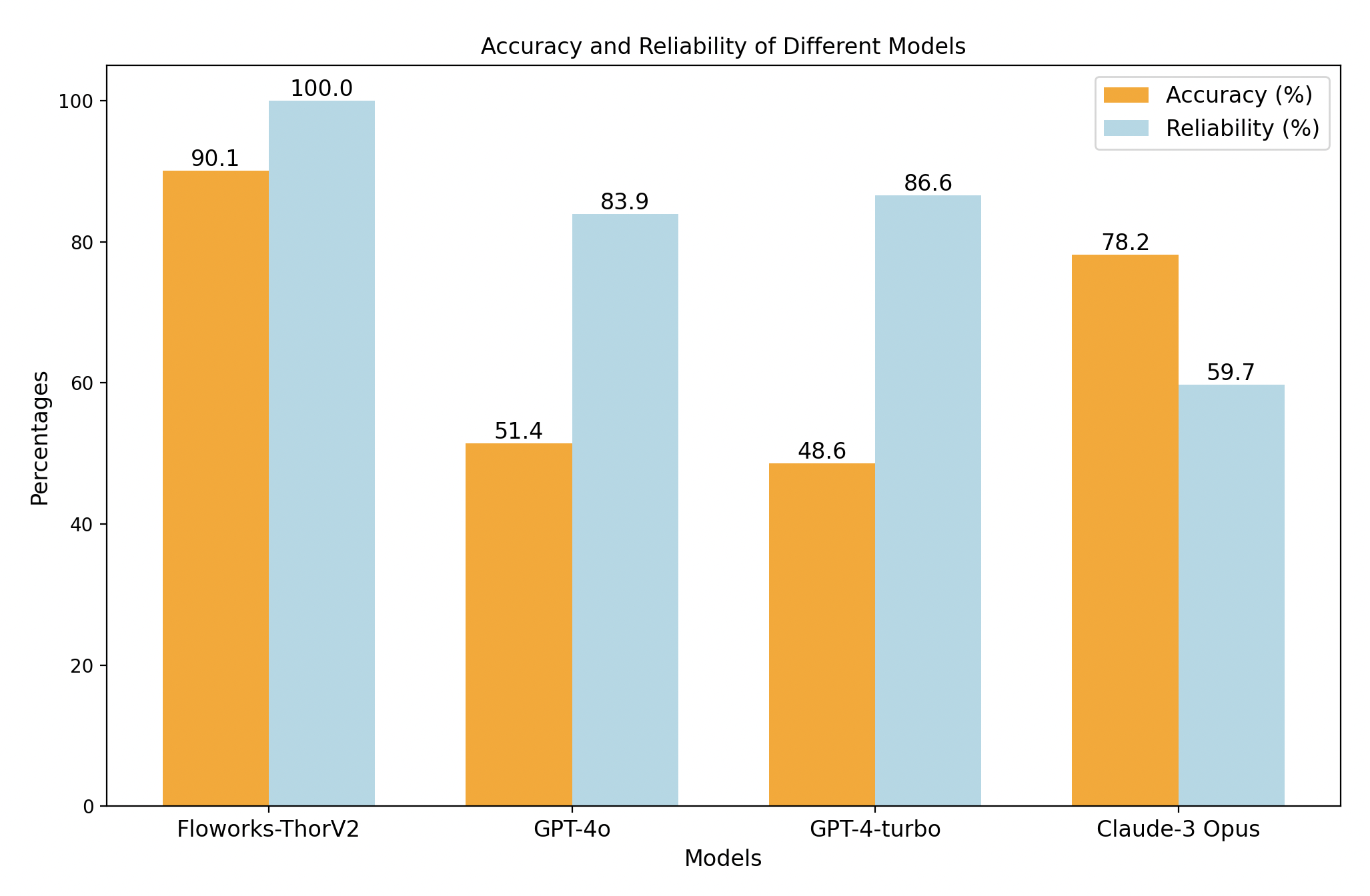}}
\caption{Single API Calling Performance - Accuracy and Reliability}
\label{fig:single_api_performance}
\end{figure}

\begin{figure}[h]
\centering
\fbox{\includegraphics[width=0.8\linewidth]{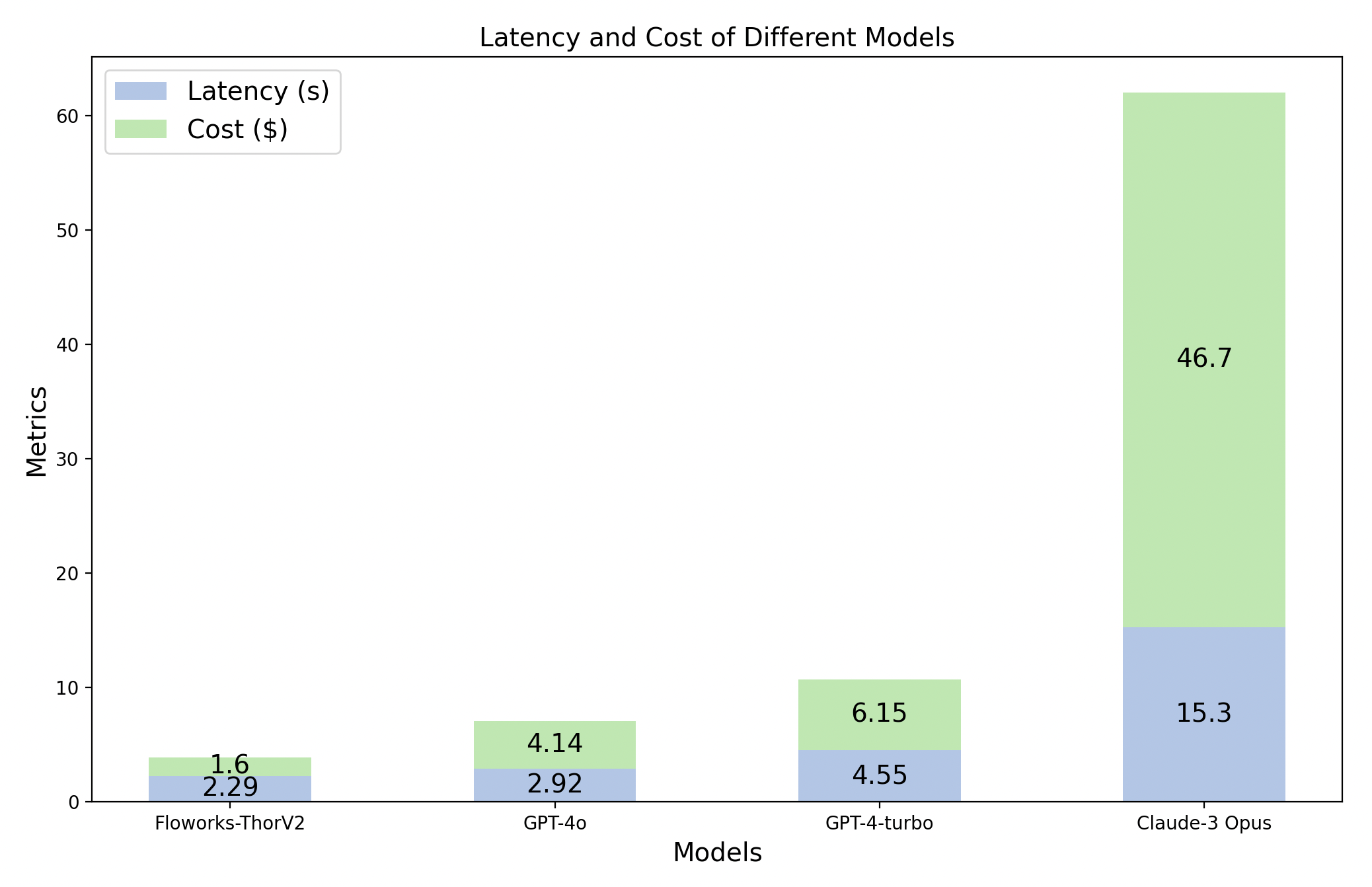}}
\caption{Single API Calling Performance - Latency and Cost}
\label{fig:single_api_performance}
\end{figure}

As shown in Table \ref{tab:single_api_results}, ThorV2 outperforms all other models across all four metrics. Notably, ThorV2 achieves a 90.1\% accuracy rate, which is significantly higher than the next best model, Claude-3 Opus, at 78.2\%. ThorV2 also demonstrates perfect reliability, ensuring consistent performance across repeated runs.


In terms of efficiency, ThorV2 exhibits the lowest latency at 2.29 seconds per query, which is 21.6\% faster than GPT-4o, the next fastest model. Cost-wise, ThorV2 is substantially more economical, with a cost per 1000 queries that is less than half of the next most cost-effective model, GPT-4o.

\subsection{Multi-API Calling Performance}
We also evaluated the models on more complex tasks requiring two API calls. The results for these multi-API calling tasks are presented in Table \ref{tab:multi_api_results}.

\begin{table}[h]
\centering
\caption{Performance metrics for multi-API calling tasks}
\label{tab:multi_api_results}
\vspace{0.7em}
\begin{tabular}{|p{0.23\textwidth}|p{0.12\textwidth}|p{0.12\textwidth}|}
\hline
Model & Accuracy & Latency (s) \\
\hline
Floworks-ThorV2 & \textbf{96.55\%} & \textbf{3.55} \\
Claude-3 Opus & 87.93\% & 36.2 \\
GPT-4o & 48.27\% & 6.83 \\
GPT-4-turbo & 51.72\% & 11.45 \\
\hline
\end{tabular}
\end{table}



In multi-API calling tasks, ThorV2 maintains its superior performance, achieving 96.55\% accuracy while keeping the latency low at 3.55 seconds. This demonstrates ThorV2's ability to handle complex, multi-step tasks efficiently and accurately.

\subsection{Scaling Laws for Latency and Cost}
\label{sec:scaling_laws}

\begin{figure}[h]
\centering
\includegraphics[width=1.0\linewidth]{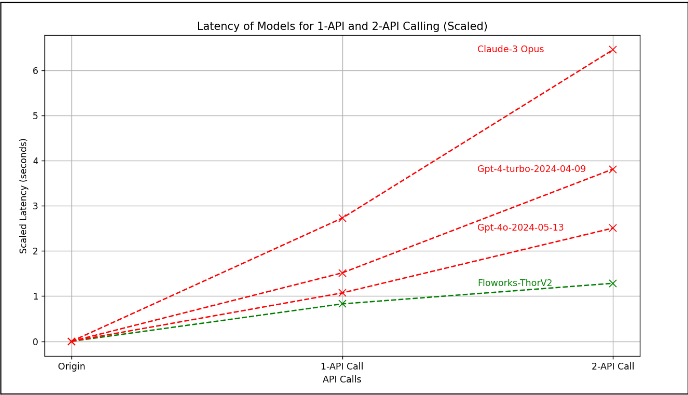}
\caption{Latency scaling for single and multi-API calling tasks}
\label{fig:latency_scaling}
\end{figure}

One of the most significant properties of ThorV2 is its superior scaling in terms of latency and cost compared to other models. Figure \ref{fig:latency_scaling} illustrates the latency scaling for single and multi-API calling tasks.

\begin{table}[h]
\centering
\caption{Latency on 1-API-calling and 2-API-calling for the four models}
\label{tab:latency_comparison}
\vspace{0.7em}
\begin{tabular}
{|p{0.23\textwidth}|p{0.15\textwidth}|p{0.15\textwidth}|}
\hline
\textbf{Model} & \textbf{1-API-calling latency} & \textbf{2-API-calling latency} \\
\hline
Floworks-ThorV2 & 2.29 & 3.55 (\textcolor{PineGreen}{+55\%}) \\
\hline
Claude-3 Opus & 15.3 & 36.2 (\textcolor{red}{+137\%}) \\
\hline
Gpt-4o-2024 & 2.92 & 6.83 (\textcolor{red}{+134\%}) \\
\hline
Gpt-4-turbo & 4.55 & 11.45 (\textcolor{red}{+152\%}) \\
\hline
\end{tabular}
\end{table}

If we compare the results of single-API-calling and 2-API-calling, we obtain an important insight. Thor-V2, due to its unique architecture and 1-shot approach to task-planning, is able to achieve a sub-linear growth in latency (2.3 seconds → 3.5 seconds), while all the other models have a super-linear growth (i.e. more than 2x). For the 2-API-call case, this is shown by the data in table 3, and represented visually in figure 13.

This reflects a more general trend – our ThorV2 has a latency and cost structure that is $N^\alpha$ with $\alpha<1$ for a complex query requiring $N$ API calls, while for a generic model the relationship is likely to be $N^\beta$ where $\beta > 1$. These trends are visualized more clearly in Figure \ref{fig:latency_scaling}. Out of the 4 models, ThorV2 is the only model that has a sublinear rate of growth in latency. This scaling advantage becomes especially important in multi-step tasks, allowing ThorV2 to maintain its performance edge even as task complexity grows.


The superior scaling of ThorV2 can be attributed to its composite planning approach (discussed in section \ref{sec:composite_planning}), which allows for more efficient handling of multi-step tasks. This approach enables ThorV2 to generate multiple API calls simultaneously, thereby minimizing the need for back-and-forth communication typically required in sequential planning.

\subsection{Query Category-wise Accuracy Breakdown}
To provide a more granular understanding of model performance, we analyzed the accuracy of each model across different categories of CRM operations: Create, Read, Update, Delete, and Associate (CRUDA).

\begin{figure}[h]
\centering
\fbox{\includegraphics[width=0.9\linewidth]{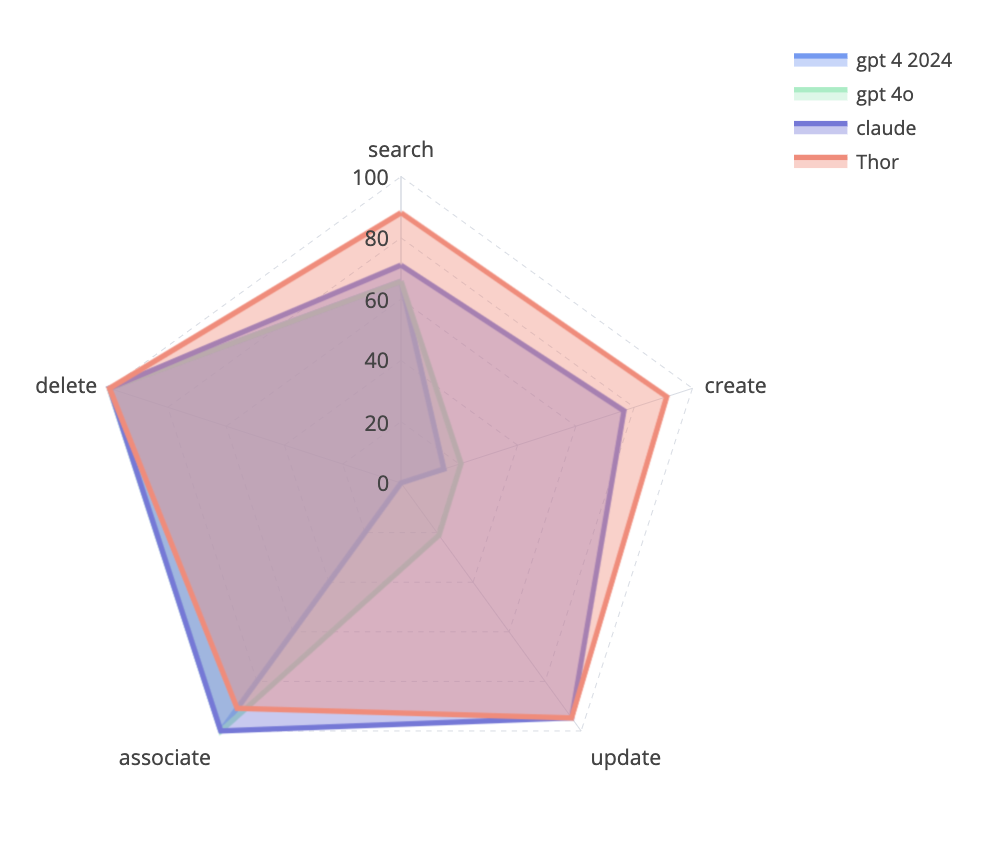}}
\caption{Accuracy Scores Across CRUDA Categories}
\label{fig:category_wise_accuracy}
\end{figure}

As illustrated in Figure \ref{fig:category_wise_accuracy}, our analysis reveals:

\begin{itemize}
    \item All models perform well on Delete and Associate operations, with high accuracy across the board.
    \item ThorV2 demonstrates a significant advantage in Create and Read operations compared to other models.
    \item For Update operations, ThorV2 and Claude-3 Opus show comparable performance, both outperforming the OpenAI models.
    \item The OpenAI models (GPT-4o and GPT-4-turbo) struggle particularly with Create and Update operations, showing markedly lower accuracy in these categories.
\end{itemize}

This breakdown highlights the strengths and weaknesses of each model across different types of CRM operations, providing valuable insights into their practical applicability in real-world scenarios.

\section{Applications}
\subsection{Implications for Business and User Experience}
The demonstrated superiority of ThorV2 over leading commercial alternatives has significant implications for businesses and end-users. The high accuracy (90.1\% for single API calls and 96.55\% for multi-API calls) translates to a more reliable user experience, reducing frustration arising from task failures or data mishandling.

The low latency of ThorV2 (2.29 seconds for single API calls and 3.55 seconds for multi-API calls) enables near-instantaneous task execution, enhancing user satisfaction and productivity. This speed advantage becomes even more pronounced in complex, multi-step tasks, where ThorV2's sublinear latency scaling shines.

ThorV2's perfect reliability score (100\%) is particularly noteworthy. This consistency allows users to develop trust in the system over time, as they can rely on it to perform the same tasks consistently. From a development perspective, this reliability makes the system much easier to engineer, maintain, and improve.

The cost-effectiveness of ThorV2 (\$1.60 per 1000 queries for single API calls) makes it an attractive option for businesses looking to implement AI assistants at scale. The significant cost savings compared to other models could be a key factor in accelerating the adoption of AI assistants across various industries.

\subsection{Potential for Generalization to Other Domains}
While our evaluation focused on HubSpot CRM operations, the principles underlying ThorV2 are domain-agnostic. The ability to perform accurate and efficient function calling is fundamental to many AI applications beyond CRM. As the AI industry transitions from mere chatbots to agents capable of taking actions in the real world, the techniques demonstrated by ThorV2 could prove invaluable across a wide range of domains.

The edge-of-domain modeling approach employed by ThorV2, which focuses on error correction rather than comprehensive upfront instructions, could be particularly useful in domains with complex, evolving APIs. This approach allows for more flexible and adaptable AI systems that can handle changes in API structures or the introduction of new functionalities with minimal retraining.

Furthermore, ThorV2's composite planning approach for multi-step tasks could be applied to various scenarios requiring sequential decision-making and action execution. This could include areas such as automated workflow management, intelligent process automation, and even robotics control systems.

\section{Limitations and Future Work}

\subsection{Limitations of Current Study}

We identify some ways in which our present results are limited. 

\begin{itemize}    
    \item \textbf{Narrow evaluation domain}: The current Benchmark is constructed exclusively on the Hubspot app. Further, it only tests moderately complex queries, up to 2 API calls in length.

    \item \textbf{Scope of latency scaling laws}: Due to time constraints, we have only tested 1-API and 2-API queries in our benchmark. However, this makes the scaling laws fragile as two data points may not seem enough to generalize. Additional data points for $N=3$ and $N=4$ API calls would make our trend lines more robust. 
        
    \item \textbf{ThorV2's implementation overhead}: ThorV2's design relies on Domain Expert Validators, which require significant time and engineering effort to develop. Moreover, this effort needs to be repeated for each new domain, i.e. each new type of software that ThorV2 must handle.
    
    \item \textbf{ThorV2's limitations in handling novel errors}: ThorV2's design relies on identifying and correcting common error patterns. Thus, novel errors or error patterns which are rare could escape detection. This could be a problem in sensitive applications, where even a 95\% accuracy rate may not be enough.

\end{itemize}

\subsection{Future Work}

Based on our discussions and the limitations noted above, the following work can be undertaken to extend our results:

\begin{itemize}   
    \item \textbf{Automated validator construction}: The main engineering effort in ThorV2 is for building the validator. In future, as LLM intelligence continues to grow, we could have a more powerful LLM (e.g. GPT-5) create validators directly.
        
    \item \textbf{Extending the Benchmark}: The "HubBench" benchmark can be extended to other domains. It can also be enhanced by including highly complex queries requiring 3 or more API calls.
    
    \item \textbf{Building Agents}: The main goal of function calling is to enable powerful AI assistants and AI agents. We can try using this approach to build agents both on software and in the real world.

\end{itemize}

\section{Conclusion}
We introduced ThorV2, a novel architecture enhancing LLM function calling capabilities. Through comprehensive evaluation on HubSpot CRM operations, we demonstrated ThorV2's superior performance over leading commercial models in accuracy, reliability, latency, and cost efficiency. Key contributions include: A new reliability metric for consistent performance assessment, Edge-of-domain modeling approach focusing on error correction, Composite planning method for efficient multi-step tasks.

ThorV2's performance suggests a promising direction for improving LLMs' practical applicability in real-world scenarios. By addressing the critical challenge of function calling, it paves the way for more capable and reliable AI assistants across various domains. As the AI industry evolves towards more agentic systems, techniques demonstrated by ThorV2 could play a crucial role in bridging the gap between theoretical capabilities and practical applications.

\section*{Acknowledgements}

This study is conducted by Floworks AI, a Y Combinator-backed (YC W23) startup. We are grateful to Y Combinator for providing compute credits on Microsoft Azure that were useful for inference. 

We would like to thank the developers of open-source function-calling models like "Gorilla" and "NexusRaven-13B" for paving the way for our work. We would also like to thank the creators of the Berkeley Function Calling Benchmark for an easy way to measure and compare existing Function-calling models. We would like to thank companies like Meta and Mistral AI for advancing the cause of open-source AI, inspiring us to open source many aspects of our model as well.

We would like to thank Soumo Deep Nandy of Floworks and Arvind Neelakantan of Meta for useful feedback and discussions.

\bibliographystyle{IEEEtran}

\bibliography{references}

\clearpage
\appendix

\section{System Prompts and Evaluation Details}

\subsection{Single Function Calling System Prompts}
\subsubsection{System Prompt for Claude-Opus}

\begin{lstlisting}[style=prompt, caption={System Prompt for Claude-Opus}]
I am hubspot owner id <owner_id>.

You are a smart function calling agent, You map all the information present in the input query to the output API call using tools provided.
You must return the output with at least two objects in the content.

You are amazingly smart and you will keep generating the output until the 'stop_reason' is 'tool_use'
and do not end the output generation when the 'stop_reason' is 'end_turn'.

you must always return the name of the tool you used to generate the function call in the output.

Do not assume any fields as required because they are present from the example in the schema.
In the input_schema provided in tools, pay attention to the required key as they are the compulsory fields and others are optional.

If the user does not provide any information, you can consider the current user (me) as the associated person.

# Rules :
- max filters per filterGroup allowed is 3.
- any timestamp should always be in the format "yyyy-MM-dd'T'HH:mm:ss.SSS'Z'"
- current time is "2024-05-05T00:00:00.000Z"
\end{lstlisting}

\subsubsection{System Prompt for GPT-4-Turbo / GPT-4o}
\begin{lstlisting}[style=prompt, caption={System Prompt for GPT-4-Turbo / GPT-4o}]
I am hubspot owner id <owner_id>.

# Rules :
- max filters per filterGroup allowed is 3.
- any timestamp should always be in the format "yyyy-MM-dd'T'HH:mm:ss.SSS'Z'"
- current time is "2024-05-05T00:00:00.000Z"
- If the timestamp is not provided, you can consider the current time as the timestamp.
- If the user is not provided, you can consider the current user (me) as the associated person.
\end{lstlisting}

\subsubsection{Prompt Example Given to All Comparison Models}

\begin{lstlisting}[style=prompt, caption={Prompt Example Given to All Comparison Models}]
Search all notes with associated deal 15860461964 (include note body, creation date, note title)

{
 "type": "tool_use",
 "id": "toolu_01BrH8mLDahkxdspT1hTBwXA",
 "name": "crm_v3_objects_notes_search_post",
 "input": {
   "after": 0,
   "filterGroups": [
     {
       "filters": [
         {
           "operator": "EQ",
           "propertyName": "associations.deal",
           "value": "15860461964"
         }
       ]
     }
   ],
   "limit": 10,
   "properties": [
     "hs_note_body",
     "hs_createdate"
   ],
   "sorts": []
 }
}
\end{lstlisting}

\subsection{Multi-API Calling System Prompts}

\subsubsection{Task Planning Prompt (Initial Step)}

\begin{lstlisting}[style=prompt, caption={Task Planning Prompt (Initial Step)}]
You are a zero-shot, chain-of-thought based logical task planner for CRM Operations. 
Given a task description, you must try to generate two steps that can be used to accurately describe how to complete the task. Any context or information provided in the task description must be used to generate the steps.

CRM has five categories of operations - Search, Create, Update, Associate, Delete. 
CRM operations are performed on objects like contacts, companies, owners, associations, lineItems, quotes, products, notes, tasks, etc.

# Rules:
- You will only respond with the sequence of steps. 
-  It may be possible that the output of one task may depend on another. 
- All the steps should be divided by a new line without indexing.
- the properties to be returned in each step are mentioned in the round brackets with the term included
- Include is not necessary for owner ID.

Associate : 
In Assign/associate queries, it should be in such a way that "Associate object A ID to object B ID." 

For example, if you are asked to "Assign the 'deal' '15810400147' to contact 'Gary'." 
To assign any object to any other object, both the objects should be IDs. So, you have to first fetch the contact ID of the object that does not have the ID. 

The sequence of output steps:

Search for the contact Gary (include id)
Associate the contact 'ID extracted from the previous task' to the deal '15810400147' 

Search : 
Searching for an object associated with another object can only be done with the Object ID except for deals/companies.
we can search for a deal/company with its name in one step; for example, Search for the deal with the name 'New Deal'.
For the rest of the objects such as contacts, owners, companies, etc., we can search for them only with their ID in a single step.
In search queries, to get the Task ID, if we have the task name/subject/title, we do it as the 'Search all tasks with subject containing'. similarly, for notes it is note body 

example query: 
Search for the deal with name 'New Deal' and owner "temp@temp.ai"

The sequence of output steps:
Search for the owner with email "temp@temp.ai" 
Search for the deal with name 'New Deal' and owner ID extracted from the previous task

Delete : 
Deleting an object can only be done with the object ID. 

Example query: 
Delete the 'company' named 'Lakka Tech Solutions' for the current user.
The sequence of output steps:
Search for the company with the name 'Lakka Tech Solutions' (include ID)
Delete the company with the given company ID extracted from the previous task

Update : 
Updating details of an object can only be done with the object ID. To update the details of an object, you first search for the object including all the details and then update the details of the object.
In update queries you can update if you have the object ID.

- Associating an object ID with owner ID is an 'UPDATE' query and not an associate query.
- For the rest of the objects it is always an associate query, example objects - contacts, companies, owners, associations, lineItems, quotes, products, notes, tasks.
- we can only update the owner of the object to a new owner ID. For example, Update the owner of the deal "12345" to "14323".

Create: 
To Create an object associated with something, you first create the object and get the object ID and then associate it with the other object.
Creating a new object associated with owner ID can be done in one step. 
But for other objects like contacts, companies, etc., we can create them in the first step and then only associate them with the object ID in the second step.

Example end query: 
Create a new 'deal' with the name 'New Deal' with owner ID '12345'. 
\end{lstlisting}

\subsubsection{Query Generation Prompt (Second Step)}

\begin{lstlisting}[style=prompt, caption={Query Generation Prompt (Second Step)}]
You are a zero-shot, single sentence Logical Query generating agent from the given information for CRM operations. 

A query can only be executed in two steps where the first step is already executed, and the second step depends on the first step. If the second task depends on the first query, your task is to generate a single sentence by reading the thoughts and the response of the first query.

All the queries are made for CRM operations.
The query will be generated to execute five different operations - Search, Create, Update, Associate, Delete.

If there are multiple IDS in the response of the first query, you can use any/all of the IDs depending on the user intent to generate the second query.

# Rules :
You must always return the output query in a "single sentence".
You generate the query in natural language instead of mathematical or technical language, without any mathematical operations
You must always frame the query in such a way that the values if they can be computed should be computed and used in the query.
Try to fetch the values from the response of the first query and use them in the second query.
all the queries must start with the category of the query.

for example, if the second thought has "increase the deal amount by 10%", then the query should be generated by reading the amount from api response and increasing it by 10%.
let's say if the amount is 1000, then the query should be "update the deal amount to 1100".

Information :
- Keep the query as simple as possible, need not include any extra information or elaborate.
- Current user ID must be replaced by 'owner 325420860' and this ID must be used in the query to be generated instead of Current user ID.
- In the response objects, id or hs_object_id are the same and can be used interchangeably.
- Associating an object ID with owner ID is an 'UPDATE' query and not an associate query.
- For the rest of the objects it is always an associate query, example objects - contacts, companies, owners, associations, lineItems, quotes, products, notes, tasks.
- we can only update the owner of the object to new owner ID. for example, Update the owner of the deal "12345" to "14323".

example queries :
- associate the deal 12345 to the contact 12345
- update the owner of the deal 12345 to 12345
- delete the deal 12345
- create a new deal with the name 'new deal' with owner 12345
- search the deal with the name 'new deal' and owner 12345
\end{lstlisting}

\end{document}